%% file: 0_main.tex
\def\year{2022}\relax
\documentclass[letterpaper]{article} 
\usepackage{aaai22}  
\usepackage{times}  
\usepackage{helvet}  
\usepackage{courier}  
\usepackage[hyphens]{url}  
\usepackage{graphicx} 
\usepackage{natbib}  
\usepackage{caption} 
\DeclareCaptionStyle{ruled}{labelfont=normalfont,labelsep=colon,strut=off} 
\usepackage{xcolor}

\usepackage{algorithm}
\usepackage{algorithmic}
\usepackage{amsfonts}
\usepackage{amsmath}
\usepackage{booktabs}
\usepackage{multirow}
\usepackage{newfloat}
\usepackage{listings}
\floatstyle{ruled}
\newfloat{listing}{tb}{lst}{}
\floatname{listing}{Listing}

\newcommand{\norm}[1]{\left\lVert#1\right\rVert}
\newtheorem{theorem}{Theorem}
\newtheorem{lemma}[theorem]{Lemma}

\title{Fast Monte-Carlo Approximation of the Attention Mechanism}

\author{
   Hyunjun Kim and JeongGil Ko
}
\affiliations{
    School of Integrated Technology, Yonsei University\\
    hyunjun.kim@yonsei.ac.kr, jeonggil.ko@yonsei.ac.kr
}

\begin{document}

\maketitle

\begin{abstract}

    We introduce Monte-Carlo Attention (MCA), a randomized approximation method for reducing the computational cost of self-attention mechanisms in Transformer architectures. MCA exploits the fact that the importance of each token in an input sequence vary with respect to their attention scores; thus, some degree of error can be tolerable when encoding tokens with low attention. Using approximate matrix multiplication, MCA applies different error bounds to encode input tokens such that those with low attention scores are computed with relaxed precision, whereas errors of salient elements are minimized. MCA can operate in parallel with other attention optimization schemes and does not require model modification. We study the theoretical error bounds and demonstrate that MCA reduces attention complexity (in FLOPS) for various Transformer models by up to 11$\times$ in GLUE benchmarks without compromising model accuracy.

\end{abstract}

\input{1_intro}
\input{2_prelim}
\input{3_mca}
\input{4_exp}

\input{5_relwork}

\input{6_sum}

\section{Acknowledgements}
This work was supported by the National Research Foundation of Korea (NRF) Grant funded by Ministry of Science and ICT (MSIT) Basic Science Research Program (2021R1A2C4002380), Information Technology Research Center (ITRC) Support Program supervised by the Institute for Information and Communications Technology Planning and Evaluation (IITP-2020-2020-0-01461), and by the Ministry of Culture, Sports and Tourism and Korea Creative Content
Agency (R2021040018).

\nocite{DBLP:journals/imwut/ParkLK21}
\nocite{DBLP:conf/aaai/ParkLK21}
\bibliography{ref}

\end{document}

%% file: 1_intro.tex
\section{Introduction}

Attention mechanisms have become a predominant modeling paradigm for state-of-the-art deep neural networks. Among many, the Transformer architecture~\cite{DBLP:conf/nips/VaswaniSPUJGKP17} is a major contribution that drove this success, which demonstrated unprecedented performance in various domains such as neural machine translation~\cite{DBLP:conf/acl/FosterVUMKFJSWB18}, question answering~\cite{DBLP:journals/jmlr/RaffelSRLNMZLL20}, image classification~\cite{DBLP:conf/nips/ParmarRVBLS19,DBLP:conf/iclr/DosovitskiyB0WZ21}, and time-series forecasting~\cite{lim2021temporal}.

A typical Transformer can be thought as a stack of self-attention layers. Self-attention can be viewed as an inductive bias with graph-like dependencies among input tokens, where important elements hold stronger influence on the final output. On an algorithmic perspective, this is a three step process: (i) computing bidirectional attention scores between input tokens, (ii) element-wise encoding of input token sequences, and (iii) computing the weighted sum of encoded sequences with respect to attention scores.

Recently, Transformer parameter sizes are ever-increasing with current trends of exploiting self-supervised pre-training using massive data. This has introduced significant computational cost when employing Transformers; thus, recent research has moved on to designing more computationally efficient attention mechanisms. 

To alleviate the computational cost of self-attention mechanisms, two mainstream approaches have been taken in previous work, where each optimize the process in different dimensions.
First is to improve the quadratic time complexity of vanilla dot-product attentions relative to the input sequence length. Since computing attention scores for an input of $n$ elements takes $\mathcal{O}(n^2)$ time and memory (typical attention being bidirectional), applying Transformers to long input sequences can be a challenge.
%
%
A number of techniques have been proposed in this dimension of optimization, such as leveraging fixed or learnable patterns to sparsify the attention matrix~\cite{DBLP:journals/corr/child19,DBLP:journals/corr/Beltagy20} and approximating attention matrices with mathematical methods~\cite{DBLP:journals/corr/Wang20, DBLP:conf/iclr/ChoromanskiLDSG21}. 

Second, there have also been efforts to reduce cost by trimming the model itself to reduce the complexity of Transformers. 
Popular approaches include quantization~\cite{DBLP:conf/aaai/ShenDYMYGMK20} and compression via knowledge distillation or pruning~\cite{DBLP:journals/corr/sanh2019,DBLP:conf/emnlp/XuZGWZ20}. Leveraging neural architecture search to remove redundant operations~\cite{DBLP:conf/ijcai/ChenLQWLDDHLZ20,DBLP:conf/nips/GuoZTC0ZH19} is also a common strategy. The efficiency benefits of these schemes are orthogonal to the first approach; thus, can be applied in parallel.

We note that existing work on reducing attention complexity generally focuses on modifying weights or changing its structure, where the optimization is essentially ``model driven''. Therefore, the performance and resource trade-off can be considered static and tightly coupled to the model or the optimization method themselves. In this work, we present a novel dimension of attention optimization, with no model weight or structure modifications, rather putting focus on the statistical characteristics of the attention matrix. Specifically, we present the \textit{Monte-Carlo Attention (MCA)}, as a flexible approach to reduce the computational cost of self-attention. MCA is motivated by the observation that attention scores are highly divergent due to the softmax operation, which prompts the question: \textit{``why should we use equal precision for encoding all input tokens?''} By suppressing computation for elements that minimally (if at all) affect the final output, we can significantly reduce the attention complexity.
%
%
MCA optimizes attention using randomized matrix multiplication approximation by configuring different error bounds for the element-wise encoding of input sequences with respect to their \textit{a priori} attention scores. We do so such that elements with weak attention scores are computed with relaxed precision, whereas errors for encoding salient elements are minimized. We discuss an optimal strategy for curtailing operational complexity with respect to the attention matrix and evaluate its theoretical error bounds. 

MCA can be applied orthogonally with already-proposed optimization approaches to complement the overall model efficiency, and holds strong advantages itself in that it (i) allows simple dynamic control of performance-resource trade-off - allowing easy adjustment to different platforms and resource availability, (ii) has predictable and parametrizable upper error bounds independent to the input length - making it applicable for various inputs, and (iii) can be used as a drop-in replacement for naive attention mechanisms by not requiring additional model preprocessing or training. 

We implement MCA with custom CUDA kernels and conduct extensive experiments using BERT~\cite{DBLP:conf/naacl/DevlinCLT19} and GLUE benchmark datasets. We also evaluate MCA's performance when applied to other common efficient Transformer variants, such as DistilBERT and Longformer to demonstrate its compatibility with other optimization schemes in parallel. Our experimental results show that MCA can effectively reduce FLOPS in attention by a large margin (up to $11\times$) with negligible accuracy drop.

%% file: 2_prelim.tex
\section{Background and Preliminaries}

We first recall to the basics of self-attention mechanisms for our problem formulation, and introduce the random sampling-based matrix approximation algorithm, which covers the background of our proposed scheme. Note that we use Python style notations for matrix indexing. For matrix $M$, its $i$-th row and column are denoted as $M[i]$, $M[:, i]$ respectively, and the element in row $i$ and column $j$ as $M[i, j]$.

\subsection{Self-Attention Mechanisms}

After the successful debut of Transformer architectures, various innovations in self-attention mechanisms have emerged. A typical self-attention mechanism in Transformers start by computing an attention matrix for the input tokens: for an input sequence $X\in\mathbb{R}^{n\times d}$ of length $n$ and feature dimension $d$, it first computes the attention matrix $A\in\mathbb{R}^{n\times n}$ based on $X$. The vanilla Transformer~\cite{DBLP:conf/nips/VaswaniSPUJGKP17} employs scaled dot-product operation to compute $A=\mathrm{softmax}(aQK^\mathsf{T})$ where $Q=XW_{q}$, $K=XW_{k}$ and $a=1/\sqrt{d}$ being the scaling factor. $W_q,W_k\in\mathbb{R}^{d\times d}$ are trainable linear projections. We note that this operation requires $\mathcal{O}(n^2)$ complexity, which induces significant overhead when long inputs are passed. Thus, researchers have tried to tackle this issue and we point interested readers to~\citealt{DBLP:journals/corr/tay2020et} for a comprehensive overview. Once the attention matrix is available, the next operation performs element-wise encoding (i.e., linear projection) of the input tokens: $H=XW$, where $W\in\mathbb{R}^{d\times d}$. Finally, the output $Y\in\mathbb{R}^{n\times d}$ is computed as product of the corresponding attention and encoded values $Y=AH$. Putting it all together, the $i$-th output element $Y[i]$ can be expressed with respect to the input element $X[i]$ as:

\begin{equation} \label{eq:gen_self_attn}
    Y[i] = \sum_{j=1}^n{A[i,j]X[j]W}
\end{equation}

Note that for simplicity, we omit the symbols for multi-head attentions, and point out that the same property holds. Asymptotically, this computation requires $\mathcal{O}(nd^2)$ time and memory. In this work we are interested in reducing the computational complexity with respect to the feature dimension $d$, which is different from approaches that optimize attention score computation for input length $n$. In many practical settings, $d$ (e.g., 768 in BERT) is much larger than $n$ and this tendency holds as Transformers scale to larger networks (e.g., $d$=3072 in Megatron-LM~\cite{shoeybi2019megatron}). Thus, we argue that optimizing for $\mathcal{O}(d^2)$ rather than $\mathcal{O}(n^2)$ can have more impact on model performance when dealing with datasets with modest sequence lengths.

\subsection{Approximating Matrix Multiplication via Monte-Carlo Sampling}
\label{sec:montecarlosampling}

Matrix multiplication approximation schemes are designed to allow formidable optimizations on large matrices by permitting some uncertainties in the final output, leading to accelerated output computation. There are several known methods for approximating matrix multiplication each with distinct statistical and computational properties. One family of algorithms is based on the low-rank approximation using truncated singular value decomposition~\cite{DBLP:conf/focs/DrineasK01,DBLP:conf/nips/DentonZBLF14,DBLP:conf/ieeehpcs/OsawaSNY17}, which well-preserves the low-rank structures of the output matrix. Unfortunately, given that they require matrix factorization as prerequisite, it is challenging to apply them in iterative processes. Fast Fourier Transform is an alternative for approximating matrix multiplication by representing the input column-row outer product as a polynomial multiplication, which can be effective when input matrices are sparse~\cite{DBLP:journals/toct/Pagh13}. In addition, kernelization-based methods demonstrated their potential in optimizing square matrix multiplications~\cite{DBLP:conf/focs/DrineasK01}. Finally, randomized algorithms~\cite{DBLP:journals/siamcomp/DrineasKM06,DBLP:journals/siamsc/Eriksson-BiqueSSWAI11}, use random sampling to estimate the original outputs with sub-sampled information. This approach does not constrain the matrix shape or contents, while offering computing acceleration.

Based on these observations, we exploit a random sampling-based approach proposed by \citeauthor{DBLP:journals/siamcomp/DrineasKM06}~\citeyear{DBLP:journals/siamcomp/DrineasKM06}, which, despite its random nature, allows for the easy control of the output matrix error bound and can be efficiently implemented on modern GPUs. Given an input matrix $A\in\mathbb{R}^{m\times k}$ and $B\in\mathbb{R}^{k\times n}$, the key idea is to view matrix multiplication as a \textit{sum of the outer products} of columns $A$ and their corresponding rows $B$. With this, we apply random sampling to a subset of these column-row pairs to compute an approximated output as the following:

\begin{equation} \label{eq:approx_mm}
    AB = \sum_{i=1}^{k}A[:,i]B[i] \approx \frac{1}{r}\sum_{i=1}^{r}\frac{1}{p(s[i])}A[:,s[i]]B[s[i]]
\end{equation}

$r\geq 1$ is the number of samples and $p(i)$ is the sampling probability of the $i$-th column-row pair. The sampling sequence $s\in\mathbb{Z}^r$ is generated via a random process $\mathbf{Pr}[s[k]=i]=p(i)$ in i.i.d. trials with replacement. The approximated output is unbiased such that $\mathbb{E}(\tilde{A}\tilde{B})=AB$ holds given sufficient samples. Since this algorithm relies on random sampling, its is important to assure that the error $\mathbb{E}(\tilde{A}\tilde{B})$, is kept low.

This algorithm allows for linear complexity reduction from $\mathcal{O}(mkn)$ to $\mathcal{O}(mrn)$ with respect to the number of samples $r$, which corresponds to the feature dimension $d$ when applied to attention mechanisms. Later, we adjust $r$ to the attention score as a way to achieve a dynamic balance between the computing overhead and output precision.

While any probability distribution can be used for $p_i$, Drineas et al. showed that applying non-uniform distribution and biasing towards higher-rank terms yields tighter upper error bounds in terms of Frobenius norm compared to a uniform distribution. Specifically, Drineas et al. proved that:
\begin{equation}  \label{eq:approx_mm_err_bound}
    \norm{AB-\tilde{A}\tilde{B}}_F = \mathcal{O}(\frac{\norm{A}_F\norm{B}_F}{\sqrt{s}}),
\end{equation}
when the $p(i)$ term is proportional to the L2 norm of each column-row pair multiplied together:

\begin{equation} \label{eq:approx_mm_sampling_prob}
    p(i) = \frac{\norm{A[:,i]}_2\norm{B[i]}_2}{\sum_{j=0}^{k}\norm{A[:,j]}_2\norm{B[j]}_2}.
\end{equation}

For details on randomized linear algebra and its algorithms, we point interested readers to \citealt{mahoney2011randomized}.


%% file: 3_mca.tex
\section{Monte-Carlo Attention}

We now present our proposed Monte-Carlo Attention (MCA), a novel approximation algorithm for reducing the complexity of self-attention mechanisms in Transformers.
MCA optimizes the element-wise encoding of input tokens (c.f., Equation \eqref{eq:gen_self_attn}) by replacing the matrix-vector product with an \textit{approximated} operation derived from random sampling. The principle idea is to apply different sampling counts for each output $Y[i]$ with respect to their attention scores to suppress overhead induced by lightly-weighted elements. Specifically, MCA can be expressed as follows:

\begin{equation} \label{eq:approx_self_attn}
    Y[i] \approx \sum_{j=1}^{n}\sum_{k=1}^{r_j}\frac{A[i,j]X[i,s_j[k]]}{r_jp(s_j[k])}W[s_j[k]]
\end{equation}

$r_i$ and $s_i$ each denote the sampling count and sampled indices for input $X[i]$, respectively, where $r_i=\mathrm{len}(s_i)$. 
While most terms in the equation are known, two need to be properly defined. The first is $p(n)$, the sampling probability for each element. Despite being able to acquire the optimal probability from Equation~\ref{eq:approx_mm_sampling_prob}, we claim that this formulation is impractical given that it requires auxiliary computation for all new input $X$. Instead, MCA disconnects dependence towards the input in computing $p(i)$, which allows the computing to be a one-time process. 
The second unknown term is $r_i$, the number of samples taken from taken from the attention matrix $A$ with respect to each input element $X[i]$. For this, MCA identifies the maximum attention score of each element towards other elements, roughly representing the minimum level of precision required for its computation. A high maximum attention score suggests that the currently evaluated element should have high precision (i.e., use more samples for encoding element $X[i]$) as it can heavily affect the output. On the contrary, a small maximum attention would indicate that the element minimally contributes to the output; thus can set a small $r_i$. Based on the theoretical upper error bounds in approximating $Y[i]$, we can derive $A$-$r$ pair relationships which guarantee robust error concentration invariant to the attention and input length.

We note that computing these parameters with respect to the error bounds is important given that MCA is essentially a randomized algorithm. The following subsections discuss how $p(i)$ and $r_i$ can be configured, and show their impact on the overall error. In a nutshell, we show that substituting Transformer attention modules to MCA, despite the randomness, guarantees tight error bounds for the attention layers.

\subsection{Formulation of Sampling Probability}

The intuition behind exploiting non-uniform distributions for sampling column-row pairs is supported by the observation that column-row pairs with relatively small norms will have minimal impact to the output. Thus, configuring a sampling probability proportional to the norm can minimize output variance. The distribution presented in Equation~\eqref{eq:approx_mm_sampling_prob} is known to be optimal in terms of minimizing the output Frobenius norm error~\cite{DBLP:conf/focs/DrineasK01}. 
%
%
Unfortunately, this formulation can be (practically) inefficient, as $p$ must be computed freshly for all incoming inputs. Furthermore, as Transformers consist of multiple self-attention layers, computing $p$ can induce an additional $\mathcal{O}(nd)$ complexity for each layer. Therefore, determining the sampling distribution only from the attention weights, and eliminating the use of input $X$, is a much more computationally-efficient strategy allowing a one-time $p$ computation. MCA borrows findings from \citealt{DBLP:journals/siamcomp/DrineasKM06} that suggests that the optimal probability can be approximated even when only one of two matrices is known. This, when applied to our domain, frees the dependency towards  the input matrix. Specifically, we compute the sampling probability by taking $W^{\mathsf{T}}W$ for approximation, which yields:

\begin{equation} \label{eq:our_sampling_prob}
    p(i) = \frac{\norm{W[i]}_2^2}{\sum_{j=0}^{d}\norm{W[j]}_2^2} = \frac{\norm{W[i]}_2^2}{\norm{W}_F^2}
\end{equation}

While this asymptotically exhibits $\mathcal{O}(d^2)$ complexity, in practice, computed results can be embedded in the model or cached to reduce the overhead. With this probability, the maximum Frobenius norm error becomes proportional to the norm of the attention weights and input, scaled by the square of the sample size.

\begin{lemma} \label{lemma:single_err_bound}
    Let $H[i]\in\mathbb{R}^{d}$ be an approximation to $X[i]W$ using $r_i\in\mathbb{Z}^+$ random samples with the probability distribution discussed in Equation~\eqref{eq:our_sampling_prob}. Then,
    \begin{equation}
        \mathbb{E}[\norm{H[i]-X[i]W}_F]\leq \frac{1}{\sqrt{r_i}}\norm{X[i]}_2\norm{W}_F.
    \end{equation}
\end{lemma}

\subsection{Determining Sample Size from Attention}

Self-attention mechanisms are designed to be bidirectional: meaning that each element in a sequence possess its own perspectives to all other elements. Hence, for any element, there exists $n$ different attention scores to describe its importance within a sequence. To determine a proper sample size $r_i$, given an element and its outbound attention scores, we take the maximum of all available scores (for each element) in defining the minimum precision. This assures high precision (i.e., more samples) for highly weighted elements and low precision (i.e., less samples) for those with low weights. When done so, based on Lemma~\ref{lemma:single_err_bound}, the mean error bound can be expressed with respect to output $Y[i]$ via Equation~\eqref{eq:gen_self_attn}: 

\begin{equation} \label{eq:error_bound}
    \mathbb{E}[\norm{\tilde{Y}[i]-Y[i]}_F]\leq \sum_{j=1}^{n}  \frac{A[i,j]}{\sqrt{r_i}}\norm{X[i]}_2\norm{W}_F.
\end{equation}

When taking $\sqrt{r_j}=\max A[:,j]$, the attention term in Equation~\eqref{eq:error_bound} can be eliminated from inequality as $\frac{A[i, j]}{\max A[:,j]}\leq 1$ for all $i$ and $j$ (assuming $A$ as non-negative). Such a strategy of taking the maximum of all attention scores can be considered conservative. Potentially, one can define more aggressive schemes by taking the mean or median of the scores. Unfortunately, analyzing the theoretical error bounds for such cases can be challenging, as error becomes dependent on the statistical characteristics of the attention matrix. We leave such optimizations for future work.

The formulation above indicates that the output error, caused from the randomness of MCA, is dependent on input sequence length $n$; thus, the error may seem to increase with long inputs. This is not desirable when bounding the error. To mitigate this, we consider an additional $n$ term in $r_i$. As a result, $r_i$ is calculated as:

\begin{equation} \label{eq:sampling_number}
    \sqrt{r_{i}}=\frac{n\cdot \max A[:, i]}{\alpha}
\end{equation}

Here, $\alpha\in(0,1)$ is the attention error coefficient, a user-controllable parameter to linearly scale error bounds. $\alpha=1$ indicates a performance-oriented configuration (i.e., reduced complexity with less samples), whereas lower $\alpha$ results in extra precision at the cost of additional computational cost. Still, we show in our evaluations that even a small $\alpha$ (e.g., 0.2) can result in significantly reduced computation.

Finally, we prove that this formulation allows robust concentration of output errors independent to the input sequence and the attention matrix statistics.

\begin{theorem} \label{theorem:y_err_bound}
    Let $\tilde{Y}[i]$ be the estimate of $Y[i]$ by equation~\eqref{eq:approx_self_attn}, and define $\beta$ as mean Euclidean norm of $X[i]$ for all $i$ (i.e., $\beta=\frac{1}{n}\sum_{i=1}^n\norm{X[i]}_2$). Assume $A$ is positive matrix for which $A[i,j]>0$ for all $i,j$. Then
    \begin{equation}
        \mathbb{E}[\norm{\tilde{Y}[i]-Y[i]}_F]\leq \alpha\beta \norm{W}_F.
    \end{equation}
    Furthermore, suppose $\delta\in (0,1)$. Then with probability at least $1-\delta$,
    \begin{equation}
        \norm{\tilde{Y}[i]-Y[i]}_F\leq \frac{\alpha\beta}{\delta} \norm{W}_F.
    \end{equation}
\end{theorem}

Therefore, we show that despite MCA's randomness, with proper $\alpha$, the overall error can still be tightly bounded; thus, achieve efficient encoding with minimal loss in accuracy.

%% file: 4_exp.tex
\section{Experiments}

\begin{table*}[t]
\begin{tabular}{@{}lllclccccccc@{}}
\toprule
\multicolumn{1}{c}{\multirow{2}{*}{Task}} & \multicolumn{2}{c}{Baseline}        &  & \multicolumn{2}{c}{$\alpha$=0.2} & \multicolumn{2}{c}{$\alpha$=0.4} & \multicolumn{2}{c}{$\alpha$=0.6} & \multicolumn{2}{c}{$\alpha$=1.0} \\ \cmidrule(l){2-12} 
\multicolumn{1}{c}{}                      & \multicolumn{1}{c}{Metric} & Result &  & Result          & FLOPS          & Result          & FLOPS          & Result          & FLOPS          & Result          & FLOPS          \\ \midrule
CoLA                                      & MC                         & 53.74  &  & 53.74$\pm$0.1   & 11.44$\times$  & 53.90$\pm$0.2   & 12.62$\times$  & 53.78$\pm$0.4   & 14.28$\times$  & 50.95$\pm$0.7   & 18.38$\times$  \\
SST-2                                     & Acc.                       & 92.43  &  & 92.26$\pm$0.0   & 5.58$\times$   & 92.04$\pm$0.0   & 6.71$\times$   & 91.22$\pm$0.0   & 8.13$\times$   & 80.66$\pm$0.1   & 12.34$\times$  \\
MRPC                                      & Acc.                       & 84.55  &  & 84.36$\pm$0.1   & 3.04$\times$   & 83.77$\pm$0.3   & 3.80$\times$   & 82.28$\pm$0.4   & 4.62$\times$   & 65.95$\pm$0.8   & 7.14$\times$   \\
                                          & F1                         & 89.41  &  & 88.96$\pm$0.0   & 3.04$\times$   & 88.86$\pm$0.2   & 3.80$\times$   & 87.69$\pm$0.3   & 4.62$\times$   & 73.50$\pm$0.7   & 7.14$\times$   \\
STS-B                                     & PC                         & 88.04  &  & 87.84$\pm$0.3   & 4.80$\times$   & 82.69$\pm$0.9   & 5.95$\times$   & 67.90$\pm$1.3   & 7.35$\times$   & 33.55$\pm$1.4   & 12.40$\times$  \\
                                          & SC                         & 87.63  &  & 87.21$\pm$0.3   & 4.80$\times$   & 81.04$\pm$0.8   & 5.95$\times$   & 67.54$\pm$1.2   & 7.35$\times$   & 28.53$\pm$1.3   & 12.40$\times$  \\
QQP                                       & Acc.                       & 90.90  &  & 90.89$\pm$0.0   & 4.76$\times$   & 90.74$\pm$0.1   & 5.73$\times$   & 89.98$\pm$0.4   & 6.89$\times$   & 78.20$\pm$0.6   & 10.30$\times$  \\
                                          & F1                         & 87.81  &  & 87.79$\pm$0.0   & 4.76$\times$   & 87.63$\pm$0.1   & 5.73$\times$   & 86.55$\pm$0.3   & 6.89$\times$   & 73.61$\pm$0.6   & 10.30$\times$  \\
MNLI                                      & Pos.                       & 83.60  &  & 83.50$\pm$0.1   & 3.75$\times$   & 82.60$\pm$0.2   & 4.76$\times$   & 78.62$\pm$0.8   & 5.88$\times$   & 71.43$\pm$1.1   & 9.61$\times$   \\
                                          & Neg.                       & 84.79  &  & 84.72$\pm$0.1   & 3.75$\times$   & 83.83$\pm$0.2   & 4.76$\times$   & 79.50$\pm$0.8   & 5.88$\times$   & 72.05$\pm$1.2   & 9.61$\times$   \\
QNLI                                      & Acc.                       & 91.54  &  & 91.47$\pm$0.0   & 3.03$\times$   & 90.24$\pm$0.0   & 3.83$\times$   & 84.33$\pm$0.1   & 4.94$\times$   & 56.43$\pm$0.1   & 10.64$\times$  \\
RTE                                       & Acc.                       & 72.56  &  & 71.68$\pm$0.2   & 2.50$\times$   & 70.39$\pm$0.4   & 3.24$\times$   & 64.72$\pm$0.8   & 4.55$\times$   & 52.65$\pm$1.1   & 10.06$\times$  \\
WNLI                                      & Acc.                       & 56.33  &  & 56.33$\pm$0.0   & 4.08$\times$   & 56.32$\pm$0.1   & 5.15$\times$   & 54.96$\pm$0.6   & 6.49$\times$   & 52.69$\pm$1.1   & 11.09$\times$  \\ \bottomrule
\end{tabular}
        \caption{\label{tab:exp-bert}FLOPS reduction and its corresponding model accuracy (with 95\% confidence intervals) when MCA-BERT is tested on the GLUE benchmark with different error bound coefficients $\alpha$. MC, PC, and SC each denote Matthews, Pearson, and Spearman correlation coefficients.}
 \end{table*}

We implement MCA via custom CUDA kernels, and measure its computation complexity and accuracy as performance metrics. We replace the multi-head attention in BERT~\cite{DBLP:conf/nips/VaswaniSPUJGKP17} with MCA, and test the performance with GLUE benchmark~\cite{DBLP:conf/emnlp/WangSMHLB18}. We also show using DistilBERT and Longformer that MCA can co-exist with existing attention optimization methods.

\subsection{Transformer Models}

\noindent\textbf{BERT} is one of the earilest applications of the Transformer architecture that demonstrated unprecedented performance in a wide spectrum of NLP tasks. We use the BERT$_{\textrm{BASE}}$ model, which encompasses 12 layers of multi-head attention with 12 heads and a feature dimension of 768.

\noindent\textbf{DistilBERT}~\cite{DBLP:journals/corr/sanh2019} is a variant of BERT which targets to reduce the computational cost by compression the Transformer's model weights using knowledge distillation~\cite{DBLP:journals/corr/HintonVD15}. DistilBERT holds the same general structure as BERT (with $\frac{1}{2}$ of the attention layers), but the token type embedding and poolers are removed from the layers. We use this model to show that MCA can coalesce with existing model compression techniques. 

\noindent\textbf{Longformer} (Beltagy et al. 2020) is a Transformer network suitable for lengthy documents by addressing the issue of complexity-blowup (i.e., $\mathcal{O}(n^2)$) with increasing input sequence lengths. Specifically, Longformers sparsify the attention matrix by employing a fixed-size window $w$ attention surrounding each input element. This constrains the complexity to $\mathcal{O}(n\times w)$, scaling linearly with input length $n$.


\subsection{Benchmark Datasets}

\noindent\textbf{GLUE Benchmark.} To consider a broad range of input data, we adopt the General Language Understanding Evaluation (GLUE) benchmark. GLUE is commonly used in evaluating the performance of language models. It consists of nine individual task sets, with two tasks for single-sentence classification (CoLA and SST-2), three tasks for semantic similarity (MPRC, QQP, and STS-B), and four natural language inference tasks (MNLI, RTE, QNLI, and WNLI).

\noindent\textbf{Document Classification.} To evaluate MCA with the Longformer model, we use the arXiv Academic Paper dataset (AAPD)~\cite{DBLP:conf/naacl/YangYDHSH16}, the IMDB review classification dataset and the Hyperpartisan News Detection (HND) dataset~\cite{DBLP:conf/semeval/KieselMSVACSP19}. AAPD consists average of 167 tokens for each entry while IMDB has entries with 300 tokens. The HND dataset includes elements with the longest item size, with an average of 705 tokens for each article.

\begin{figure}[t]
    \centering
    \includegraphics[width=0.5\textwidth]{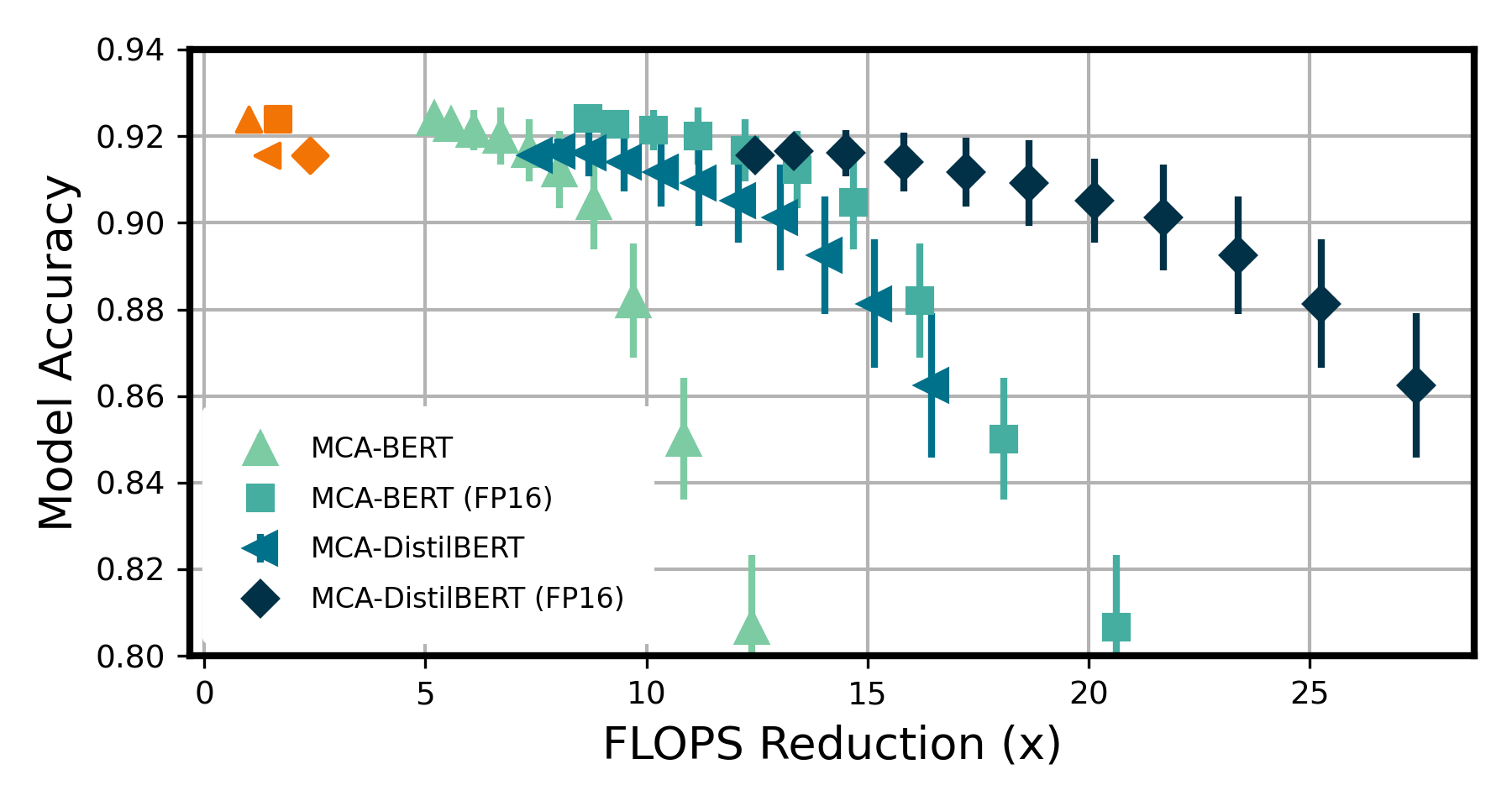}
    \caption{Mean accuracy and FLOPS trade-off for MCA-BERT and MCA-DistilBERT, along with FP16 quantized versions (SST-2 dataset). The orange plots on the top-left show the accuracy-FLOPS relationship for original BERT and DistilBERT (without MCA applied).}
    \label{fig:tradeoff}
\end{figure}

\begin{figure}[t]
    \centering
    \includegraphics[width=0.5\textwidth]{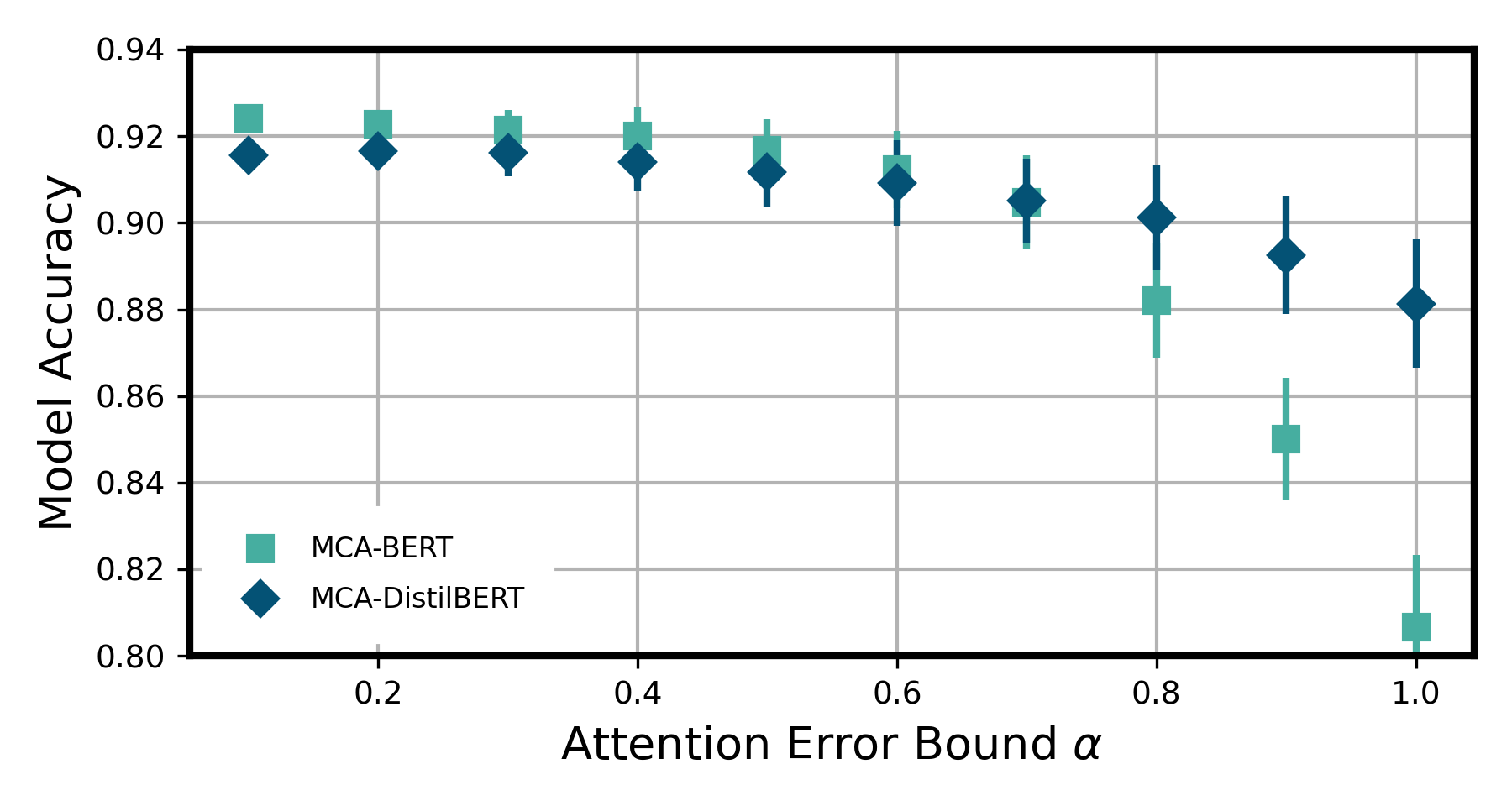}
    \caption{Impact of attention error bound $\alpha$ on model accuracy of MCA-BERT and MCA-DistilBERT (SST-2 dataset). The vertical bars represents 95$\%$ confidence interval. }
    \label{fig:err}
\end{figure}

\subsection{Implementation and Experimental Setup}

MCA is implemented as a C++ PyTorch extension using custom CUDA kernels. Our implementation replaces the self-attention in Transformer models, and can be configured whether to run in {\tt regular mode} (i.e., act as a normal self-attention) or {\tt approximation mode} (i.e., activate MCA). We employ AWS \texttt{p3.2xlarge} instances for the experiment, which serves one NVIDIA V100 GPU and use the Huggingface library for implementations and download pre-trained weights for BERT, DistilBERT, and Longformer. Readers can refer to Appendix 1 for reproducibility details.

\subsection{FLOPS Reduction and Accuracy Impact}

Using the GLUE benchmark, we first evaluate the performance of MCA when applied to BERT. Specifically, we measure the Floating Operations Per Second (FLOPS) of the MCA-BERT attention operation and compare with when MCA is not applied. Note that we measure only the FLOPS for the attention (i.e., $AXW$) and exclude other Transformer operations (e.g., embedding layer, classification heads) as they will be consistent. We also report the full model accuracy to observe the complexity-accuracy trade-off. We compute 128 samples for each task (different random seeds) and report the mean with 95\% confidence interval. Given that $\alpha$ is an adjustable parameter that determines the attention error bound tightness, we evaluate for $\alpha =$ 0.2, 0.4, 0.6, and 1.0.

Table~\ref{tab:exp-bert} summarizes our results. Here, MCA shows a significant FLOPS reduction for all GLUE tasks. Using $\alpha=0.2$, MCA exhibits 4.64$\times$ FLOPS reduction (mean of all tests), with negligible accuracy loss. With increasing $\alpha$, we see further model complexity reduction. For example, with $\alpha=0.4$, we observe $5.72\times$ average FLOPS reduction with $\leq 1\%$ accuracy loss for all cases. A more generous $\alpha\geq 0.6$, offers even more drastic improvements in complexity reduction. However, this comes at the cost of accuracy drop, some quite noticeable, suggesting that $\alpha$ should be configured with respect to the target task and dataset. 

One important observation is that the FLOPS reduction, while maintaining consistent trends, varies with respect to the dataset. For example, at $\alpha=0.2$, the CoLA task shows a $11.44\times$ FLOPS reduction compared to $2.50\times$ of RTE. Given that MCA reduces complexity by exploiting smaller sample sizes for elements with low attention scores, we conjecture that differences can rise from the fact that the attention matrix from CoLA is sparser than that of RTE. Generally, binary classification tasks (e.g., CoLA and SST-2) exhibit higher reduction rates while semantic similarity tasks and NLI tasks show modest reduction rates.

Another point to note is that previously proposed schemes that compress or prune the model require the tuning of many different hyperparameters. On the contrary we argue that the effort needed for per-task $\alpha$ adjustment in MCA is minimal, as this is the only hyperparameter to adjust when applying on different datasets or computing environments.


\noindent\textbf{Performance-Accuracy Trade-off.}
Figure~\ref{fig:tradeoff} visualizes the required FLOPS and model accuracy relationships of BERT, DistilBERT, MCA-BERT, and MCA-DistilBERT using the SST-2 dataset (top-left orange plots are cases with MCA not applied). We also present results for the corresponding 16-bit floating point (FP16) quantized models of each case, respectively. As the results show, MCA results in 40-60\% FLOPS reduction with (nearly) the same accuracy and as much as 170\% reduction with small accuracy loss. Nevertheless, if the FLOPS drops below a tipping point, MCA shows noticeable loss in accuracy, showing a logarithmetic tradeoff relationship. Figure~\ref{fig:tradeoff} also shows that performance trends of MCA holds for models with quantized weights as well.

\noindent\textbf{Attention Error vs. Model Error.} Figure~\ref{fig:err} presents the impact of the attention error bound $\alpha$ on the model accuracy for BERT and DistilBERT with MCA applied when using the SST-2 dataset. As the plots show, with increasing $\alpha$ (higher errors at the attention), the overall model accuracy shows a gradually decreasing pattern. Nevertheless, configuring $\alpha$ to 0.4, or even 0.6 will only minimally impact the overall model accuracy. Comprehensively with the results reported in Table~\ref{tab:exp-bert} we can see that despite the small loss in accuracy, the computational complexity reduction gain is significant.

\begin{table*}[t]
    \begin{tabular}{@{}lllclccccccc@{}}
\toprule
\multicolumn{1}{c}{\multirow{2}{*}{Task}} & \multicolumn{2}{c}{Baseline}        &  & \multicolumn{2}{c}{$\alpha$=0.2} & \multicolumn{2}{c}{$\alpha$=0.4} & \multicolumn{2}{c}{$\alpha$=0.6} & \multicolumn{2}{c}{$\alpha$=1.0} \\ \cmidrule(l){2-12} 
\multicolumn{1}{c}{}                      & \multicolumn{1}{c}{Metric} & Result &  & Result          & FLOPS   & Result          & FLOPS   & Result          & FLOPS   & Result          & FLOPS   \\ \midrule
CoLA                                      & MC                         & 56.85  &  & 56.49$\pm$0.1   & 11.36$\times$  & 55.69$\pm$0.3   & 12.33$\times$  & 54.29$\pm$0.3   & 13.75$\times$  & 50.08$\pm$0.6   & 17.30$\times$  \\
SST-2                                     & Acc.                       & 91.51  &  & 91.50$\pm$0.0   & 5.60$\times$   & 91.41$\pm$0.0   & 6.64$\times$   & 90.92$\pm$0.0   & 7.84$\times$   & 88.20$\pm$0.1   & 10.58$\times$  \\
MRPC                                      & Acc.                       & 85.78  &  & 85.26$\pm$0.1   & 2.98$\times$   & 83.83$\pm$0.2   & 3.73$\times$   & 78.59$\pm$0.4   & 4.54$\times$   & 64.76$\pm$0.5   & 7.35$\times$   \\
                                          & F1                         & 90.26  &  & 89.91$\pm$0.1   & 2.98$\times$   & 88.99$\pm$0.2   & 3.73$\times$   & 85.66$\pm$0.2   & 4.54$\times$   & 76.04$\pm$0.5   & 7.35$\times$   \\
STS-B                                     & PC                         & 88.39  &  & 88.29$\pm$0.2   & 4.65$\times$   & 79.73$\pm$0.3   & 5.91$\times$   & 63.11$\pm$0.4   & 7.21$\times$   & 30.12$\pm$0.7   & 12.33$\times$  \\
                                          & SC                         & 87.56  &  & 86.94$\pm$0.2   & 4.65$\times$   & 78.24$\pm$0.3   & 5.91$\times$   & 62.40$\pm$0.4   & 7.21$\times$   & 29.02$\pm$0.6   & 12.33$\times$  \\
QQP                                       & Acc.                       & 88.24  &  & 88.11$\pm$0.0   & 4.75$\times$   & 87.79$\pm$0.2   & 5.85$\times$   & 86.77$\pm$0.5   & 7.14$\times$   & 72.24$\pm$0.9   & 10.41$\times$  \\
                                          & F1                         & 85.48  &  & 84.99$\pm$0.0   & 4.75$\times$   & 83.26$\pm$0.2   & 5.85$\times$   & 82.95$\pm$0.8   & 7.14$\times$   & 70.02$\pm$1.3   & 10.41$\times$  \\
MNLI                                      & Pos.                       & 82.50  &  & 82.37$\pm$0.1   & 3.62$\times$   & 81.09$\pm$0.4   & 4.31$\times$   & 77.70$\pm$0.6   & 5.09$\times$   & 68.11$\pm$0.8   & 7.40$\times$   \\
                                          & Neg.                       & 83.83  &  & 83.66$\pm$0.1   & 3.62$\times$   & 82.41$\pm$0.4   & 4.31$\times$   & 79.26$\pm$0.6   & 5.09$\times$   & 71.25$\pm$0.8   & 7.40$\times$   \\
QNLI                                      & Acc.                       & 88.48  &  & 88.31$\pm$0.0   & 2.86$\times$   & 87.06$\pm$0.0   & 3.51$\times$   & 83.19$\pm$0.1   & 4.28$\times$   & 62.89$\pm$0.2   & 7.43$\times$   \\
RTE                                       & Acc.                       & 65.70  &  & 65.22$\pm$0.3   & 2.33$\times$   & 64.43$\pm$0.6   & 2.91$\times$   & 60.35$\pm$0.8   & 3.64$\times$   & 52.79$\pm$0.9   & 6.47$\times$   \\
WNLI                                      & Acc.                       & 56.33  &  & 54.40$\pm$0.6   & 3.97$\times$   & 52.55$\pm$1.4   & 5.10$\times$   & 52.77$\pm$1.5   & 6.56$\times$   & 52.24$\pm$1.9   & 10.55$\times$  \\ \bottomrule
\end{tabular}
        \caption{\label{tab:exp-distilbert} Experiment results on MCA-DistilBERT. Experimental configurations are kept same as Table~\ref{tab:exp-bert}.}
 \end{table*}

\subsection{Integration with Compressed Transformers}

We next evaluate the impact of applying MCA on already compressed Transformer models. For this, we perform the same experiment as Table~\ref{tab:exp-bert} on DistilBERT, which only possess $40\%$ of BERT's original parameter size. We present results in Table~\ref{tab:exp-distilbert}. Overall, results from MCA-DistilBERT mostly agree with those seen for MCA-BERT in terms of FLOPS reduction and accuracy. For more a detailed view, we look back on Figure~\ref{fig:tradeoff} which plots a finer-grained FLOPS-accuracy relationship for both MCA-applied models (using SST-2), which shows that MCA effectively reduces the complexity even for already compressed models.

From Figures~\ref{fig:tradeoff} and~\ref{fig:err}, we see that the accuracy drop with increasing $\alpha$ is more gradual for MCA-DistilBERT compared to MCA-BERT. Since, MCA is a randomized approach, while tightly bounded, a small amount of error accumulates with more attention layers to process. With MCA-DistilBERT having only $\frac{1}{2}$ attention layers of MCA-BERT, we conjecture that MCA-DistilBERT benefits by having less errors accumulated.


\begin{table*}[t]
    \begin{tabular}{@{}lllclccccccc@{}}
\toprule
\multicolumn{1}{c}{\multirow{2}{*}{Task}} & \multicolumn{2}{c}{Baseline} &  & \multicolumn{2}{c}{$\alpha$=0.2} & \multicolumn{2}{c}{$\alpha$=0.4} & \multicolumn{2}{c}{$\alpha$=0.6} & \multicolumn{2}{c}{$\alpha$=1.0} \\ \cmidrule(l){2-12} 
\multicolumn{1}{c}{}                      & Metric        & Result       &  & Result          & FLOPS   & Result          & FLOPS   & Result          & FLOPS   & Result          & FLOPS   \\ \midrule
AAPD                                      & Acc.          & 74.81        &  & 74.70$\pm$0.1   & 3.44$\times$   & 72.21$\pm$0.2   & 4.45$\times$   & 69.68$\pm$0.2   & 6.22$\times$   & 66.87$\pm$0.3   & 9.83$\times$   \\
                                          & F1            & 71.39        &  & 71.02$\pm$0.1   & 3.44$\times$   & 68.33$\pm$0.2   & 4.45$\times$   & 65.35$\pm$0.3   & 6.22$\times$   & 64.14$\pm$0.3   & 9.83$\times$   \\
HND                                       & Acc.          & 80.33        &  & 80.18$\pm$0.0   & 5.32$\times$   & 78.73$\pm$0.1   & 6.81$\times$   & 74.23$\pm$0.3   & 7.93$\times$   & 70.36$\pm$0.5   & 11.81$\times$  \\
                                          & F1            & 76.57        &  & 76.32$\pm$0.0   & 5.32$\times$   & 74.53$\pm$0.1   & 6.81$\times$   & 72.94$\pm$0.3   & 7.93$\times$   & 68.53$\pm$0.4   & 11.30$\times$  \\
IMDB                                      & Acc.          & 89.11        &  & 89.05$\pm$0.0   & 3.67$\times$   & 84.10$\pm$0.1   & 5.06$\times$   & 81.17$\pm$0.3   & 7.50$\times$   & 75.11$\pm$0.4   & 11.30$\times$  \\ \bottomrule
\end{tabular}
        \caption{\label{tab:exp-longformer} Experiment results on MCA-Longformer. A local attention with window size of 256 is used for the Longformer model.}
 \end{table*}

\subsection{Integration with Sparse Attention Patterns}

To empirically validate the coalesced performance of MCA and Transformers with sparse attention mechanisms, we integrate MCA to the Longformer model and test its performance on the document classification task datasets. We configure the Longformer with a window size of 256, and apply global attention for the \texttt{CLS} token.

As results in Table~\ref{tab:exp-longformer} show, for all datasets, MCA achieved FLOPS reduction by a factor of at least $3\times$ with negligible error (for tight error bounds). The tendency of FLOPS-accuracy trade-off for the Longformer experiments also agree with those from prior observations. Comprehensively, these results suggest that MCA can be generally applied with different Transformer architecture variants.

%% file: 5_relwork.tex
\section{Related Work}

A number of previous work target to alleviate the computational cost of self-attention mechanisms, mostly in two orthogonal ways: (i) by optimizing for long input sequences, and (ii) by proposing new Transformer designed that are less resource demanding. We introduce examples below. 

\subsection{Efficient Attention for Long Sequences}

The scaled dot-product attention in a vanilla Transformer induces 
significant overhead when dealing with lengthy data. 
Thus, a number of previous work have focused on optimizing the evaluation algorithms for attention scores as a way to efficiencize Transformers~\cite{DBLP:conf/icml/TayBMJZZ21}.

\noindent \textbf{Fixed Sparse Patterns.} Some early efforts exploited sparse attention with fixed patterns, which reduce the attention complexity proportional to a target sparsity. As an example, blocked attention~\cite{DBLP:conf/emnlp/QiuMLYW020,DBLP:conf/nips/ParmarRVBLS19} groups input elements in blocks for block-wise attention evaluation. More recently, strided-pattern attention has been proposed~\cite{DBLP:journals/corr/child19,DBLP:journals/corr/Beltagy20}. These work limit attention evaluation to fixed-sized windows and evaluates only the neighboring elements.

\noindent \textbf{Learnable Sparse Patterns.} 
Methods that learn the access pattern have also been proposed. Reformer~\cite{DBLP:conf/iclr/KitaevKL20} replaces dot-product attention with locally-sensitive hashing, which clusters input tokens based on hash similarity and filters prominent tokens. Online K-means clustering~\cite{DBLP:journals/tacl/RoySVG21} can be used as a similar concept by learning the input sparsity patterns.

\noindent \textbf{Low-Rank Approximations.} 
Some work directly approximate attention scores by assuming a low-rank structured attention matrix~\cite{DBLP:journals/corr/Wang20}. More recently, Performer~\cite{DBLP:conf/iclr/ChoromanskiLDSG21} exploited kernelization for approximation, which does not rely on input sparsity patterns or low-rankness.

\subsection{Low-cost Transformers}
There have also been efforts to downscale the time complexity of Transformers themselves as we detail below.

\noindent \textbf{Model Compression.}
Pruning redundant information in Transformer weights~\cite{DBLP:conf/rep4nlp/GordonDA20,DBLP:conf/emnlp/WangWL20} and less-important heads~\cite{DBLP:conf/nips/MichelLN19} have shown good performance albeit its complicated post-training procedure. Another approach is to compress the model via knowledge distillation~\cite{DBLP:journals/corr/Beltagy20}, allowing flexible model sizes at the cost of (potentially heavy) re-training.

\noindent \textbf{Weight Quantization.}
Along with FP16 quantization, more aggressive quantization schemes, such as 8-bit integers~\cite{DBLP:conf/nips/ZafrirBIW19} or even down to 4- or 2-bits~\cite{DBLP:conf/aaai/ShenDYMYGMK20}, have been applied to Transformers. While this line of research does not often involve heavy post-processing (e.g., additional model training), such low precision schemes, in practice, often require hardware accelerators to maximize the desired efficiency.

\noindent \textbf{Neural Architecture Search (NAS).} 
Employing NAS downsizes Transformer sizes by identifying more efficient model architectures. This can be very effective when targeting a specific hardware architecture~\cite{DBLP:conf/acl/WangWLCZGH20} or when meeting target latency goals~\cite{DBLP:conf/nips/GuoZTC0ZH19}. 

\subsection{Positioning MCA with Previous Work}

Despite some limitations, we acknowledge that active research in both perspectives are meaningful. As discussed, MCA takes a different approach, by exploiting the statistical characteristics of attention matrices. Such an approach can be considered to be orthogonal to the previously proposed improvements, and we showed through our evaluations that MCA can indeed operate in parallel with other computation acceleration techniques for Transformers.

%% file: 6_sum.tex
\section{Conclusion}

This work presented the Monte-Carlo Attention (MCA) mechanism, a randomized approximation method for reducing the complexity of self-attention mechanisms in Transformers. MCA is designed under the philosophy that not all attention elements need to be treated with equal weight. Those with heavy attention should be computed with high precision, whereas the weaker, which give minimal impact to the final output, can use lower precision. For this MCA employs random sampling-based matrix-vector product operations to perform element-wise encoding of input tokens, where each element is allocated different amounts of samples with respect to their attention score. We evaluated the model complexity reduction performance (in terms of FLOPS) and the model accuracy using BERT (and its variations) with GLUE benchmark datasets. Our results suggest that MCA is an effective alternative to the attention operators used in today's Transformer networks by reducing the FLOPS by up to $11\times$ with negligible (near-zero) accuracy drop. Furthermore, we showed that MCA can be applied in parallel with widely-used approaches to efficiencize Transformers, while still maintaining its performance benefits.